\documentclass[12pt]{article}

\usepackage{verbatim}
\usepackage{fancyvrb}
\usepackage{varwidth}
\usepackage[section]{placeins}
\usepackage{placeins}
\usepackage[utf8]{inputenc}
\usepackage{kotex}
\usepackage{microtype} 
\usepackage{PRIMEarxiv}
\usepackage[utf8]{inputenc} 
\usepackage[T1]{fontenc}    
\usepackage{hyperref}       
\usepackage{url}            
\usepackage{booktabs}       
\usepackage{amsfonts}       
\usepackage{nicefrac}       
\usepackage{microtype}      
\usepackage{lipsum}
\usepackage{fancyhdr}       
\usepackage{graphicx}       
\graphicspath{{media/}}     
\usepackage{multirow}

\usepackage{cite}
\usepackage{amsmath,amssymb,amsfonts}
\usepackage{mathtools}
\usepackage{graphicx}
\usepackage{textcomp}
\usepackage{algorithm}
\usepackage{algpseudocode}
\usepackage{algorithmicx} 
\usepackage{booktabs}
\usepackage[utf8]{inputenc}
\usepackage{subcaption}

\usepackage{fancyhdr} 
\usepackage{lipsum}   

\title{Knowledge-Aware Iterative Retrieval for
Multi-Agent Systems
}

\author{
  Seyoung Song \\
  \texttt{\{ssong\}@gatech.edu} \\
}

\begin{document}



\maketitle 
\thispagestyle{fancy}

\begin{abstract}
We introduce a novel large language model (LLM)-driven agent framework, which iteratively refines queries and filters contextual evidence by leveraging dynamically evolving knowledge. A defining feature of the system is its decoupling of external sources from an internal knowledge cache that is progressively updated to guide both query generation and evidence selection. This design mitigates bias-reinforcement loops and enables dynamic, trackable search exploration paths, thereby optimizing the trade-off between exploring diverse information and maintaining accuracy through autonomous agent decision-making. Our approach is evaluated on a broad range of open-domain question answering benchmarks, including multi-step tasks that mirror real-world scenarios where integrating information from multiple sources is critical, especially given the vulnerabilities of LLMs that lack explicit reasoning or planning capabilities. The results show that the proposed system not only outperforms single-step baselines regardless of task difficulty but also, compared to conventional iterative retrieval methods, demonstrates pronounced advantages in complex tasks through precise evidence-based reasoning and enhanced efficiency. The proposed system supports both competitive and collaborative sharing of updated context, enabling multi-agent extension. The benefits of multi-agent configurations become especially prominent as task difficulty increases. The number of convergence steps scales with task difficulty, suggesting cost-effective scalability. When agents utilize a lightweight LLM, the system achieves results that are comparable or superior to those obtained with heavier counterparts, indicating that the agent system architecture enhances reasoning capabilities independently of the underlying LLM's reasoning abilities. This also points to the potential for further optimization of agent collaboration structures. 
\end{abstract}

\section{Introduction}
Large Language Models (LLMs) are probabilistic language generation models that do not incorporate explicit reasoning systems or logical planning modules. Consequently, in tasks that require synthesizing information over multiple steps, the reasoning performed at each stage is not clearly delineated, and intermediate reasoning occurs implicitly, making the process susceptible to errors. Furthermore, the difficulty of rigorously validating each step exacerbates the accumulation of errors throughout the overall process.

To overcome these challenges, it is often necessary to retrieve external knowledge that compensates for the inherent limitations of LLMs, especially in real-world scenarios. Approaches such as Retrieval Augmented Generation (RAG) play a significant role by acquiring information not contained within the model in real time, thereby enabling more precise responses.

Multi-step question answering (QA) is a representative challenge that demands both high precision in intermediate reasoning and the integration of diverse information. It not only exposes the limitations of LLMs but has also emerged as an important benchmark for real-world problems that seek to transcend these limitations.

In this context, we propose Knowledge-Aware Iterative Retrieval for Multi-Agent Systems, a retrieval optimization system that employs an agent-based framework. It iteratively optimizes search queries through agent-guided knowledge accumulation, with a focus on query refinement, the iterative process of modifying or enhancing an initial query to improve search results. The system dynamically optimizes search queries using LLM-based agents and operates through the following core stages:
\begin{enumerate} \item \textbf{Query Planning:} Agents leverage accumulated knowledge to propose or refine search queries while avoiding redundant searches and ensuring that unresolved sub-goals are addressed. \item \textbf{Knowledge Extraction:} Relevant passages from retrieved documents are distilled into verified facts and identification of unresolved gaps, which are then used to update each agent’s knowledge base. \item \textbf{Contextual Filtering:} Utilizing the accumulated knowledge, the system filters out extraneous or inconsistent text to retain only the most pertinent document segments, thereby reducing reasoning overload and mitigating potential confusion or hallucination. \end{enumerate}
If information gaps persist, the system reiterates the cycle of query planning, knowledge extraction, and contextual filtering. Moreover, in a multi-agent configuration, agents can operate in parallel by sharing knowledge or refined context, thereby more efficiently exploring the search space.

The novelty of the proposed system lies in its ability to balance the conflicting objectives of collecting a diverse range of information while ensuring accuracy. Conventional approaches in multi-step question answering generally do not refine intermediate search results because the process of information filtering, such as determining relevance, is a multi-step challenge that LLMs struggle to address. In contrast, the proposed method optimizes the trade-off between the conflicting objectives through agent-based autonomous decision-making.
By enabling targeted gap detection and re-querying mechanisms supported by dynamically updated internal knowledge, the system mitigates LLM biases toward specific keywords or contexts, thereby facilitating self-correction. Moreover, search queries are progressively refined, and both the queries and the accumulated knowledge are maintained in a structured form, enhancing traceability and enabling the identification of errors at specific stages.

During the processes of knowledge extraction, contextual filtering, and query execution, we leverage an LLM as an optimization tool. Additionally, by sharing refined knowledge or processed documents, the system can be scaled to a multi-agent configuration, facilitating more efficient exploration of the search space.

The proposed method has been extensively evaluated across a diverse array of open-domain QA benchmarks, including those that require multi-step reasoning. Our findings are as follows:

\begin{itemize} \item The agent-based retrieval system outperforms single-step retrieval baselines (e.g., naive RAG) regardless of the task. In our evaluation, we considered both retrieval effectiveness and downstream QA performance. 

\item Compared to conventional iterative retrieval approaches based on static search strategies, the proposed method offers several advantages. It reduces computational cost through systematic query expansion, improves scalability, and enables precise evidence-based reasoning. As tasks require more compositional reasoning, the proposed approach prioritizes preserving precision over indiscriminate context expansion, resulting in efficiency gains over conventional methods.

\item When multi-agent extensions are applied, the performance boost compared to a single agent becomes particularly evident for more complex tasks, indicating that the optimal resolution strategy is task-dependent. However, the number of convergence steps in the agent system scales with task difficulty, ensuring that efficiency is maintained across varying levels of complexity. In competitive multi-agent extensions, implicit role differentiation among agents emerges, thereby enhancing the proposed system’s systematic gap resolution mechanism, which plays a key role in addressing challenging tasks.

\item Agent scaling results indicate that performance gains do not increase linearly with the number of agents, and that LLMs reported to have superior reasoning capabilities are not necessarily better suited for collaborative frameworks. For example, a 2-agent \texttt{GPT-4o-mini} configuration achieved cost-effective, optimal results compared to configurations using \texttt{GPT-4o}. This finding suggests that the benefits of the agent architecture can enhance problem-solving capabilities independently of the underlying LLM's reasoning abilities, and underscores the potential for further optimization of agent collaboration mechanisms.
\end{itemize}

\section{Related Work}
\subsection{Retrieval-augmented generation (RAG)}
Retrieval-Augmented Generation (RAG) combines traditional information retrieval with generative models to handle open-domain QA tasks, improving factual accuracy by retrieving relevant documents from external sources and generating answers. However, simple implementation of RAG, where often one model instance is responsible for query understanding, retrieval, and answer generation, often face significant challenges due to cascading errors, a phenomenon where errors made early in the retrieval or reasoning process propagate through later stages, compounding mistakes and reducing overall system accuracy. \cite{yoran2024makingretrievalaugmentedlanguagemodels} empirically demonstrate that retrieval-augmented language models are highly sensitive to the relevance of the retrieved context: while relevant passages enhance performance, irrelevant ones can lead to cascading errors, particularly in multi-hop reasoning scenarios. To address this, they propose a modular, black-box solution that leverages a natural language inference (NLI) model to filter out irrelevant passages without altering the underlying LLM's parameters.

\subsection{Multi-Step Question Answering}  
Multi-step question answering requires models to systematically integrate information across sequential reasoning steps, a task particularly challenging for conventional LLMs. While humans naturally decompose complex queries into modular sub-problems, LLMs often struggle with implicit error propagation due to their lack of explicit reasoning mechanisms. This limitation has spurred the development of specialized benchmarks to rigorously evaluate multi-step reasoning capabilities.  

For instance, \cite{trivedi2022musiquemultihopquestionssinglehop} introduces dependency-chained sub-questions where each step's resolution hinges on the prior step's correctness, effectively eliminating shortcut solutions. Similarly, \cite{ho2020constructingmultihopqadataset} provides structured reasoning paths to trace whether models genuinely perform multi-hop inference rather than answer surface matching. These benchmarks underscore the critical need for explicit intermediate verification, a gap addressed by our proposed knowledge-aware iterative retrieval framework.  

\subsection{Multi-Step Retrieval Optimization}
In response to the limitations of conventional single-step retrieval methods such as RAG, recent research has increasingly focused on iterative retrieval approaches. For instance, \cite{trivedi2023interleavingretrievalchainofthoughtreasoning} interleaves retrieval with chain-of-thought reasoning, refining queries at each iteration based on intermediate inferences from a generator such as an LLM. Although these methods tend to show high accuracy in downstream QA benchmarks, they incur high computational costs as iterations repeat and risk exceeding context-window limits since all intermediate results are fed back into the model.

To address these limitations, an adaptive retrieval approach was proposed in \cite{jeong2024adaptiveraglearningadaptretrievalaugmented} that employs a dedicated classifier to determine when multi-step retrieval is needed. By triggering multi-step retrieval based on task difficulty, this method aims to mitigate the computational overhead associated with repeated retrieval and reasoning cycles. However, this method is still fundamentally based on an iterative retrieval approach, employing a separate model, specifically a classifier trained to classify task difficulty.

\section{Proposed Approach}
\label{sec:Retrieval_Optimization}
The proposed approach employs iterative retrieval but distinguishes itself through the following key mechanisms:

The system enables diversity in search exploration via targeted query formulation. Unlike conventional methods that accumulate all intermediate reasoning outputs from a separate generative model (e.g., chain-of-thought), our approach maintains a dedicated knowledge base derived from LLM-generated outputs while decoupling it from query formulation. By explicitly designing queries to isolate necessary information at each step, the system preserves a transparent reasoning trajectory, critical for real-world applications requiring verifiable intermediate steps. Furthermore, this separation mitigates model-inherent biases and provides a foundation for multi-agent extensions through shared knowledge cross-pollination.

In addition, the system ensures accuracy through dynamic context filtering. While existing methods naively retain all intermediate contexts, leading to computational overload and LLM confusion, our approach bounds context size by selectively filtering irrelevant or redundant information. Although technically challenging (requiring multi-step summarization and extraction), the filtering process leverages an agent-curated knowledge base independent of external documents, thereby reducing susceptibility to hallucinations.

Finally, the system optimizes the trade-off between diversity and accuracy through agent-based autonomous decision-making and multi-agent cross-validation. By dynamically prioritizing either exploration (diverse search paths) or exploitation (evidence convergence) based on task requirements, the framework achieves robustness against conflicting objectives. This optimization not only improves processing efficiency but also enables scalable multi-agent extensions while maintaining performance across complexity levels.


\subsection{Core Components}

The proposed agent-based knowledge enhanced retrieval system implements an iterative inference process as detailed in Algorithm~\ref{alg:agent_inference}. The architecture optimizes the accuracy--diversity trade-off through interconnected components:

\begin{itemize}
  \item \textbf{Knowledge Update Mechanism}:
At each step, the agent independently makes decisions based on the LLM, including query formulation, knowledge updating, and document filtering. To facilitate this, the system maintains two dynamic memory structures: 
    \begin{itemize}
 \item \textit{What is Known}:  $\mathcal{K}_t = \{k_1,\dots,k_n\}$ where each $k_i$ represents verified facts.
        \item \textit{What is Required}: $\mathcal{R}_t = \{r_1,\dots,r_m\}$ where each $r_j$ denotes an unresolved information gap, which is a specific piece of missing knowledge required to resolve the overarching query.
 \end{itemize}

At each step, the system evaluates the relevance of content relative to the query and extracts the core information from the source documents, which may include summarization or editorial refinement. In doing so, the system structurally decouples the relevance assessment from the filtering process, ensuring that only information directly pertinent to the inquiry is retained (\textit{What is Known}).

The system dynamically defines and tracks the unresolved information gaps that must be addressed (\textit{What is Required}). This enables the autonomous decomposition of user inputs into components that are more targeted and progressively refined.

The knowledge structure undergoes continuous refinement through interactions with the external environment, such as document repositories. As the system collects new documents via search tools to address unresolved information gaps in \(\mathcal{R}_t\), it extracts and integrates relevant information into \(\mathcal{K}_t\) (verified facts) while updating \(\mathcal{R}_t\) to reflect remaining gaps. This process establishes a closed feedback loop: updated knowledge in \(\mathcal{K}_t\) directly informs subsequent query formulation to avoid redundant searches, while revised gaps in \(\mathcal{R}_t\) guide the prioritization of unexplored information needs. By iteratively aligning retrieved evidence with both verified facts and unresolved requirements, the system dynamically adapts its search strategy, ensuring contextually informed retrieval that balances the exploration of new search paths with the exploitation of confirmed evidence. This iterative refinement mechanism enables progressive convergence toward resolving the overarching query while maintaining computational efficiency.


\item \textbf{Query Planning}:  
  The agent dynamically formulates search queries by referring to:  
    \begin{itemize}
        \item Current state of \textit{What is Known} ($\mathcal{K}_t$)  
        \item Unresolved information gaps in \textit{What is Required} ($\mathcal{R}_t$)  
        \item \textit{Query history} ($\mathcal{Q}_t$)  to prevent redundancy  
    \end{itemize}
While prioritizing the resolution of unresolved sub-problems, the agent also autonomously identifies missing information gaps through its internal reasoning. This dual mechanism, which is grounded in both explicit goals and agent-derived hypotheses, enables progressive decomposition of complex objectives and exploration of diverse search paths, thereby enhancing adaptability in dynamic environments.

\item \textbf{Contextual Filtering}:  
The system employs a dual-stage filtering process that leverages the dynamically updated knowledge \(\mathcal{K}_t\). As described in the Knowledge Update Mechanism, potentially conflicting information in \(\mathcal{K}_t\) is fact-checked against new documents \(\mathcal{D}_t\) (integrating or discarding data based on consistency checks). Then, the system refines \(\mathcal{D}_t\) at the passage level:
\[
\mathcal{D}_{filtered} = \bigcup_{d \in \mathcal{D}} \{ s \in d \mid \phi(s, \mathcal{K}_t) > \tau \},
\]
where \(\phi\) is a semantic matching function (e.g., an LLM-based approach for measuring textual similarity) that computes the relevance between a text segment \(s\) and the current knowledge \(\mathcal{K}_t\). Since \(\mathcal{K}_t\) is iteratively updated based on new documents and inference steps, each filtering operation adapts to the latest state of verified facts.

\end{itemize}

\subsection{Optimization Strategy}
\label{sec:opt_strategy}
\noindent
The proposed system adheres to the core principles of agent autonomy and goal-directed behavior through two interconnected mechanisms: an iterative decision cycle and dynamic goal management. These ensure self-directed adaptation to evolving information needs while maintaining alignment with overarching objectives.

\medskip
\begin{itemize}
    \item \textbf{Autonomous Decision Cycle}:
     The agent executes the following steps without external intervention, demonstrating autonomy through closed-loop reasoning:
    \begin{enumerate}
        \item \textit{Query Generation}:
            \[
            q_{t+1} = f_{LLM}(\,q_{t}, \mathcal{D}_{t}\;|\;\mathcal{K}_t,\mathcal{R}_t, \mathcal{Q}_{1:t}\,),
            \]
            where the agent synthesizes queries using verified facts ($\mathcal{K}_t$), unresolved gaps ($\mathcal{R}_t$), and query history ($\mathcal{Q}_{1:t}$) to avoid redundancy.
        \item \textit{Document Retrieval}:
            \[
            \mathcal{D}_{\text{add},t+1} = \text{Retrieve}\bigl(q_{t+1}\bigr),
            \]
            expanding evidence while preserving exploration diversity.
            
        \item \textit{Knowledge Update}:
            \[
            \mathcal{K}_{t+1},\;\mathcal{R}_{t+1} 
            = g_{LLM}\bigl(\mathcal{D}_t \;\big|\; \mathcal{K}_{t},\;\mathcal{R}_{t}\bigr),
            \]
             refining verified facts and unresolved gaps through agent-driven validation.
            
        \item \textit{Filter Document}:
            \[
            \mathcal{D}_{t+1} 
            = h_{LLM}\Bigl(
                \bigl(\mathcal{D}_{t} \cup \mathcal{D}_{\text{add},t+1}\bigr) 
                \;\Big|\;\mathcal{K}_{t+1}
              \Bigr).
            \]
             retaining only knowledge-aligned passages to minimize reasoning noise.
    \end{enumerate}
    
    \item \textbf{Goal Management}:
    The system balances dual objectives through adaptive termination:
    \begin{itemize}
        \item Primary objective: 
Minimize unresolved gaps: $\min|\mathcal{R}_t|$, while maximizing evidence relevance: $\max \text{Relevance}(\mathcal{D}_t)$. 
        \item Dynamic termination condition:

\begin{equation}
\label{eq:termination}
\text{End}_t = \mathbb{I}\Bigl[
    \mathcal{R}_t = \emptyset \;\lor\; \text{Sufficiency}(M_t, \mathcal{D}_t) \ge \tau 
\Bigr]
\end{equation}
where \(\mathcal{R}_t = \emptyset\) indicates that all required information gaps have been resolved; \(M_t\) is the moderator (optional, activated when external validation is available) that assesses whether the current refined context \(\mathcal{D}_t\) sufficiently answers the user query; and \(\tau\) is a moderator-determined relevance threshold.

    \end{itemize}    
\end{itemize}

\noindent
The inherent conflicts between accuracy and diversity poses a critical challenge: while diverse search exploration fosters creative connections, it risks introducing unverified information. Conversely, strict adherence to verified facts limits novel insights. To resolve this trade-off, the proposed system integrates the following optimization mechanisms for autonomous balancing of context preservation (minimizing hallucination) and information diversity (maximizing exploration):
\medskip
\begin{itemize}

  \item \textbf{Progressive Diversity Promotion}:
\begin{equation}
  \label{eq:div_promotion}
  \begin{split}
    q_{t+1} = \arg\max_{q'} \Bigl[ 
      &\alpha \,\text{Div}(q' \mid \mathcal{Q}_{1:t}) \\
      &+ (1 - \alpha)\,\text{Rel}(q' \mid \mathcal{R}_t) \Bigr]
  \end{split}
\end{equation}
    \noindent

where \(q'\) denotes a candidate query generated during the query formulation process. \( \text{Div}(q' \mid Q_{1:t}) \) measures how different \(q'\) is from the previously attempted queries \(Q_{1:t}\), thereby promoting information diversity. Meanwhile, \( \text{Rel}(q' \mid R_t) \) quantifies how relevant \(q'\) is to the current requirements \(R_t\). The weight \(\alpha\) is dynamically adjusted based on the agent’s assessment of query diversity needs versus task priority.

  \item \textbf{Context-Preserving Filtering}:

\begin{equation}
    \label{eq:context_filtering}
    \begin{split}
        \text{Accept}(d') 
        = \mathbb{I}\Bigl[ 
          &\text{ExactMatch}(d', d) = 1 
          \;\land\; \\
          &\max_{\mathclap{k \in \mathcal{K}_t}} \text{Rel}(d', k) > \tau_2 \Bigr]
    \end{split}
\end{equation}

\noindent
where 
\(\text{Rel}(d', d)\) captures how closely the refined context \(d'\) remains aligned to the original document \(d\) (ensuring that the context is preserved), and 

\[
\text{ExactMatch}(d', d) = 
\begin{cases} 
1, & \text{if } d' \subseteq d,\\[6pt]
0, & \text{otherwise}.
\end{cases}
\]

\noindent
which guarantees that \(d'\) does not include any modified passages (e.g., it prevents transformations such as \(p_1 \rightarrow x_1\)). Additionally, \(\max_{k \in \mathcal{K}_t}\text{Rel}(d', k)\) measures how well \(d'\) aligns with the knowledge base \(\mathcal{K}_t\), enforcing factual consistency. The function \(\mathbb{I}[\cdot]\) is an indicator function, and \(\tau_2\) is an agent-determined relevance threshold.

\item \textbf{Knowledge Update:}
\begin{itemize}
    \item \textbf{Binary Fact-Checking:}  

\begin{equation}
\label{eq:flag_or_discard}
\begin{aligned}
\text{FlagOrDiscard}(s)
&= \mathbb{I}\Bigl[
   \nexists\, k \in \mathcal{K}_t \text{ such that } \\
&\quad \text{Supported}(s, k) = 1
\Bigr]
\end{aligned}
\end{equation}
where \(\text{Supported}(s, k)\) is a binary function that returns 1 if segment \(s\) is directly supported by a verified fact \(k \in \mathcal{K}_t\), and 0 otherwise, ensuring that any segment not supported by the current knowledge is flagged or discarded.

\item \textbf{Dynamic Knowledge Integration:}

\begin{equation}
\label{eq:knowledge_update}
\begin{aligned}
\mathcal{K}_{t+1} 
&= \mathcal{K}_t \cup \Bigl\{
   s \,\Big|\,
   \text{Supported}(s, d) = 1 \\
&\quad \land \,
   \text{Relevant}\bigl(s, \mathcal{R}_t\bigr) = 1 
\Bigr\}
\end{aligned}
\end{equation}

where \(\text{Relevant}(s, \mathcal{R}_t)\) returns 1 if segment \(s\) addresses the unresolved information gaps in \(\mathcal{R}_t\), modeling the iterative refinement process by which new evidence is integrated into the knowledge base, ensuring that only supported and relevant information is added.
\end{itemize}
\medskip
Together, these mechanisms help to reduce the risk of unfounded claims and ensure that the system's knowledge base is continuously updated in a contextually informed manner.

\end{itemize}
\medskip
In summary, each mechanism addresses a distinct subproblem:

\begin{enumerate} \item \textbf{Progressive Diversity Promotion} balances the need for exploring new or alternative search paths against their relevance to task priority in order to fill unresolved information gaps. \item \textbf{Context-Preserving Filtering} ensures that any refined context remains consistent with both verified knowledge and source documents. 
\item \textbf{Knowledge Update} ensures factual consistency by flagging or discarding any segments that are not supported by the current knowledge base or do not meet the moderator's criteria. Then, it dynamically integrates relevant and verified evidence into the knowledge base, thereby continuously refining its understanding.
\end{enumerate}

\medskip
By decoupling these mechanisms, the system optimizes the accuracy–diversity trade-off.   

\begin{algorithm}[t]
\caption{Knowledge-Aware Agent Retrieval Algorithm}
\label{alg:agent_inference}

\textbf{Require:} 
Refinement Step, Knowledge Update Step, Filtering Step, 
Search Tool \( T \), (External) Router \( R_{\mathrm{ext}} \)

\begin{itemize}
    \item \textbf{Status Cache Definitions:}
    \begin{itemize}
        \item \(\mathbf{K} \subseteq \mathcal{I}\) (Known Information)
        \item \(\mathbf{R} \subseteq \mathcal{I}\) (Required Information)
        \item \(\mathbf{Q} \subseteq \Sigma^*\) (Query History)
        \item \(\mathbf{D} \subseteq \mathcal{D}\) (Refined Context)
    \end{itemize}
\end{itemize}

\begin{itemize}
    \item \textbf{Input:} User question \( x \)
    \item \textbf{Output:} Refined context \(\mathbf{D}\) sufficient to answer \( x \)
\end{itemize}

\begin{algorithmic}[1]
    \State \textbf{Initialize}: 
    \(\mathbf{K} \leftarrow \emptyset,\;
      \mathbf{R} \leftarrow \emptyset,\;
      \mathbf{Q} \leftarrow \emptyset,\;
      \mathbf{D} \leftarrow \emptyset\).
      
    \State \textbf{Check External Router} \( R_{\mathrm{ext}} \) for retrieval need.
    \If {Retrieve == Yes}
        \State \textbf{(Initial Retrieval)}: Call \( T(x) \).
        \State \(\quad\rightarrow\) Obtain rephrased queries \(\mathbf{q}_0\) and passages \(\mathbf{d}_0\).
        
        \State \textbf{[Knowledge Update]}: 
            Update \(\mathbf{K}, \mathbf{R}\) using \(\mathbf{d}_0\).
            
        \State \textbf{[Filter Context]}:
            Derive refined context \(\mathbf{D}\) from \(\mathbf{d}_0\) using \(\mathbf{K}\), \(\mathbf{R}\).
            
        \State \(\mathbf{Q} \gets \mathbf{Q} \cup \{\mathbf{q}_0\}\); 
               \(\mathbf{D} \gets \mathbf{d}_0\);
        \State \( i \gets 1 \)
        
        \Repeat
            \State \textbf{[Refine]}: 
                Call \( T(\mathbf{q}_{i-1}) \). 
                Obtain \(\mathbf{q}_i\), \(\mathbf{d}_i\).
            \State \(\mathbf{Q} \gets \mathbf{Q} \cup \{\mathbf{q}_i\}\);
                   \(\mathbf{D} \gets \mathbf{D} \cup \mathbf{d}_i\);
                   
            \State \textbf{[Knowledge Update]}:
                Update \(\mathbf{K}, \mathbf{R}\) using \(\mathbf{D}\).
                
            \State \textbf{[Filter Context]}:
                Re-filter \(\mathbf{D}\) with the updated \(\mathbf{K}, \mathbf{R}\).
                
            \State \textbf{[Check Knowledge]}:
                If \(\mathbf{R} = \emptyset\), terminate. 
                Else, \(i \gets i + 1\).
                
        \Until {termination condition is met}
    \Else
        \State \textbf{Terminate the process}.
    \EndIf
    \State \Return \(\mathbf{D}\)
\end{algorithmic}
\end{algorithm}
 
\subsection{Multi-Agent Extension}

\label{sec:multi_agent_extension}

\noindent
Although described for a single agent, the framework’s modular design and explicit state management inherently support multi-agent extensions.

In particular, the shared knowledge base or context can serve as a collaborative blackboard for multi-agent communication. A dedicated goal management agent could assign information gaps (\(\mathcal{R}_t\)) to specialized agents, while individual agents could still employ distinct strategies. 

\medskip
\noindent
\textbf{Competition--Collaboration Mechanism.}
The proposed framework accommodates both competitive and collaborative multi-agent mechanisms:
\begin{itemize}
    \item \textbf{Competition Example:}
Agents maintain separate knowledge bases ($\mathcal{K}_t^{(i)}$) and generate proposals independently. The system selects the optimal proposal via a utility metric (Equation~\ref{eq:competition_model}), updating the shared refined context. No direct coordination occurs; agents compete through proposal quality.  
    \item \textbf{Collaboration Example:}
    Agents share a unified knowledge base (\(\mathcal{K}_t\)). For instance, Agent 1 resolves \(\mathcal{R}_t^{(1)}\), updating \(\mathcal{K}_t\), which then informs Agent 2’s resolution of \(\mathcal{R}_t^{(2)}\). This interdependence ensures global consistency.  
\end{itemize}  

\noindent

\begin{figure}[t!]
  \centering
  \includegraphics[width=0.8\linewidth, keepaspectratio]{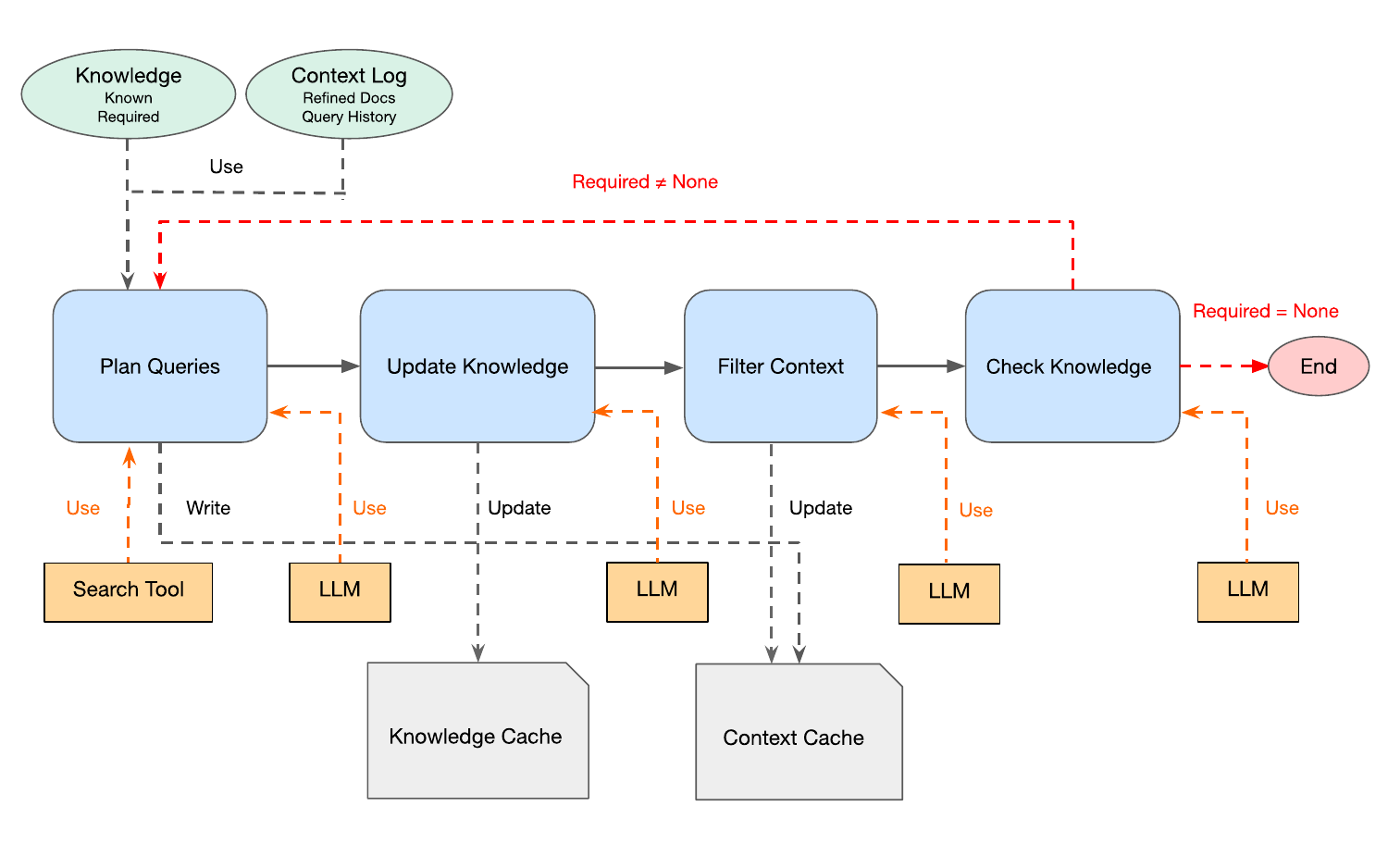}
  \caption{
    Overview of the single-agent architecture featuring a dynamic knowledge management and decision cycle.  
    Iterative refinement balances accuracy and diversity through modular optimization.  
  }
  \label{fig:single_agent}
\end{figure}

\begin{figure}[t!]
  \centering
  \begin{subfigure}[t]{0.48\linewidth} 
    \centering
    \includegraphics[width=\linewidth, keepaspectratio]{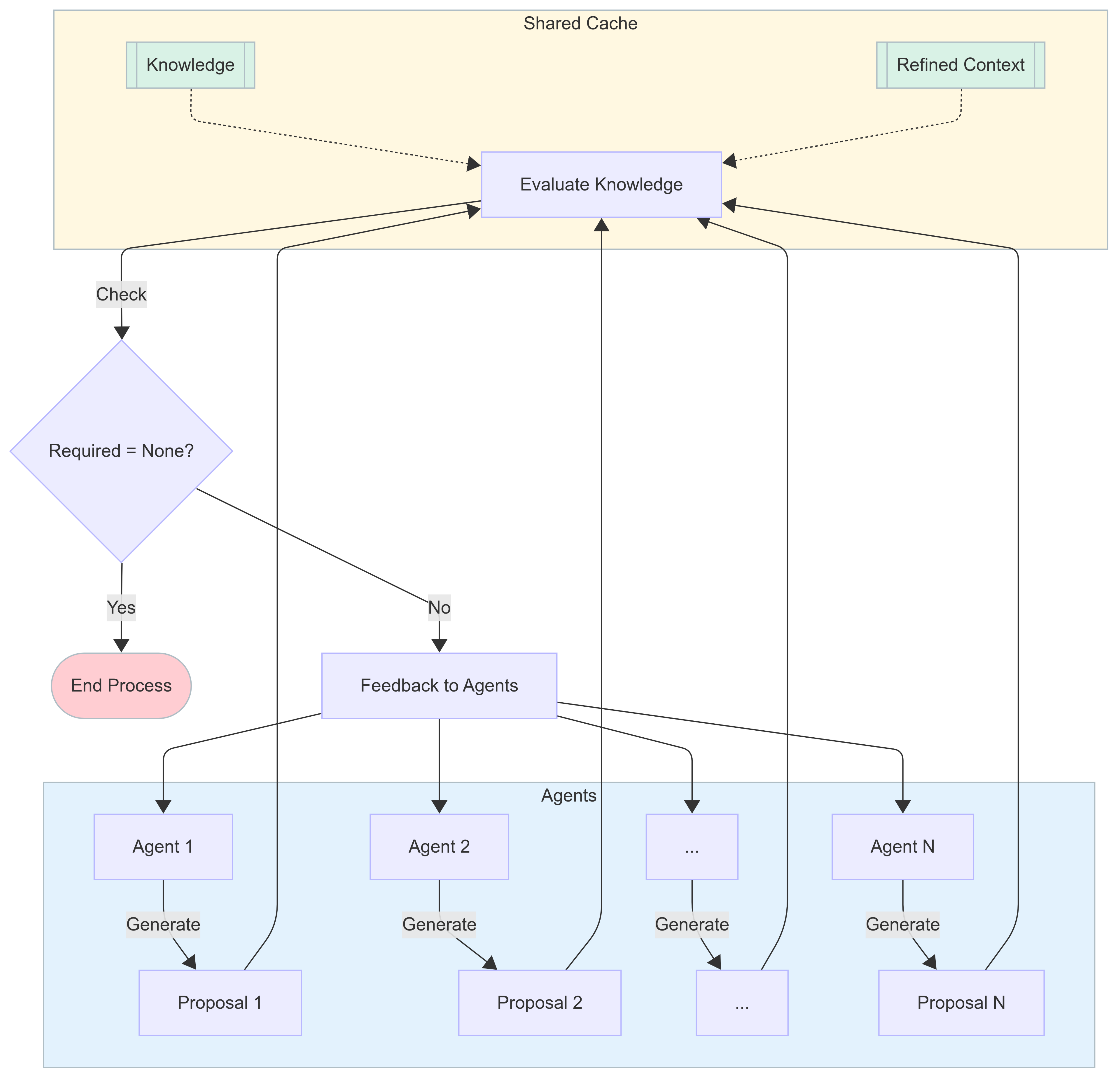}
    \caption{Collaborative model with a shared knowledge base}
  \end{subfigure}
  \hfill
  \begin{subfigure}[t]{0.48\linewidth} 
    \centering
    \includegraphics[width=\linewidth, keepaspectratio]{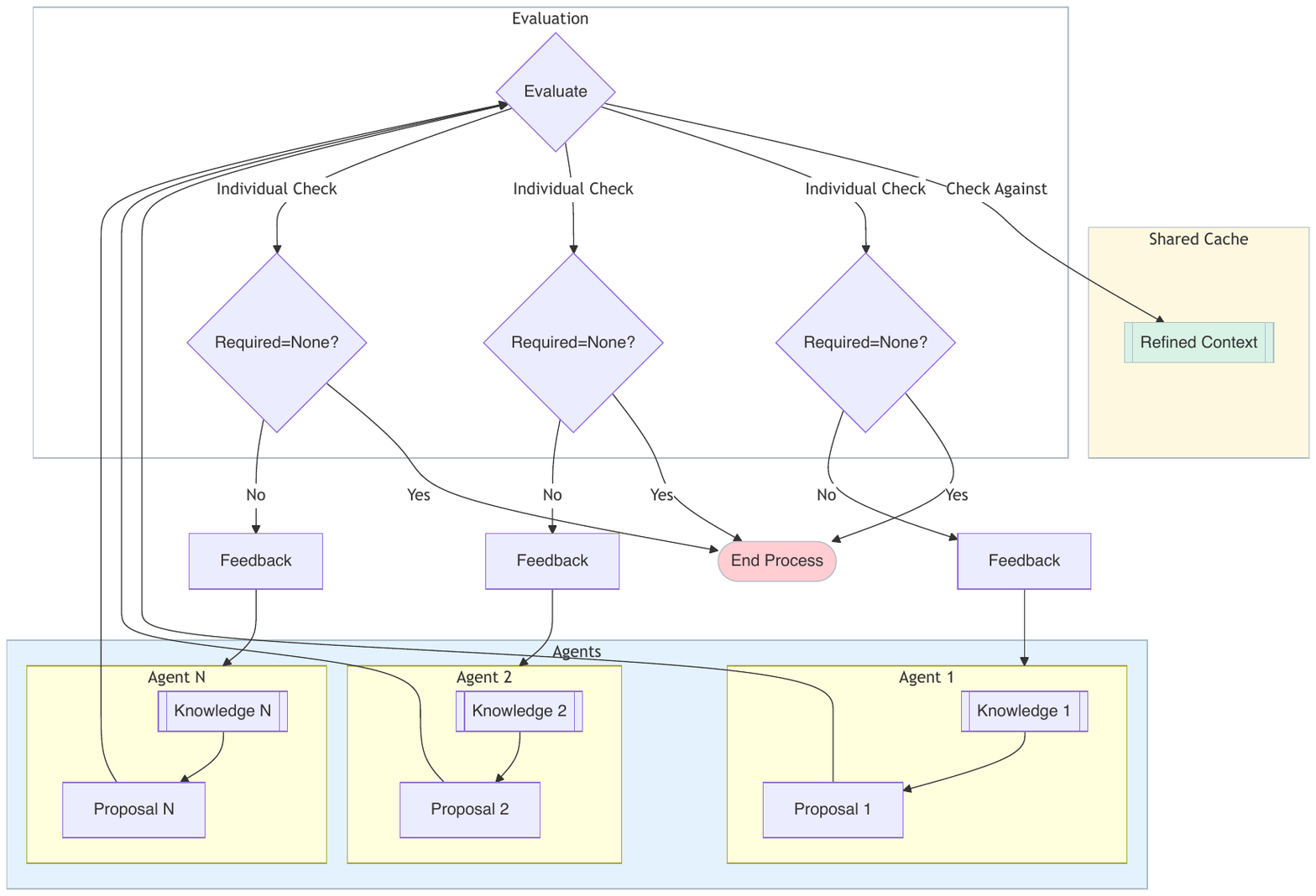}
    \caption{Competitive model with agent-specific knowledge}
  \end{subfigure}
  
  \caption{
    Multi-agent architectures: (\textbf{Left}) Collaborative model, where agents share and update a common knowledge repository.  
    (\textbf{Right}) Competitive model, where agents maintain individual knowledge to reduce coordination overhead.
  }
  \label{fig:multi_agent}
\end{figure}

As illustrated in \textbf{Figure~\ref{fig:single_agent}} and \textbf{Figure~\ref{fig:multi_agent}}, the single-agent system sequentially executes all phases, while the multi-agent system distributes these phases across agents that are connected via a shared knowledge base and context.

\medskip
\noindent
\textbf{Competition-Driven Implementation.}
We adopt a competition based strategy to avoid the overhead of collaborative message passing. Agents independently propose solutions, with the system prioritizing proposals that maximally resolve \(\mathcal{R}_t\). Formally, let \(\mathcal{A} = \{A_1, A_2, \dots, A_n\}\) be the set of agents. Each agent \(A_i\) generates a candidate requirement set \(\mathcal{R}_{t+1}^{(i)}\) based on its local \(\mathcal{K}_t^{(i)}\). The system evaluates all proposals and prioritizes the agent achieving the minimal unresolved requirements:

\noindent
\begin{equation}
    \label{eq:competition_model}
    \mathcal{R}_{t+1}^*
    =
    \arg\min_{\mathcal{R}_{t+1}^{(i)}}
    \left|\mathcal{R}_{t+1}^{(i)}\right|,
    \quad
\end{equation}

\noindent
Termination occurs immediately if any agent resolves all gaps (\(\mathcal{R}_t^{(i)} = \emptyset\)), ensuring efficiency without global consensus.

\medskip
\noindent
\section{Experiments}
\label{sec:Experiments}

We now describe how we measured the performance of the proposed system against against baseline configurations.

\subsection{Experimental Settings}

\subsubsection{Tasks and Datasets}

We evaluate two core aspects of the system:  
\begin{itemize}  
\item \textbf{Retrieval Effectiveness}:  
  Measures the system’s ability to comprehensively retrieve all ground-truth documents, including those critical for multi-step reasoning.  
\item \textbf{Question Answering}:  
  A separate generator, which operates as an independent pipeline, receives the final retrieved output and produces a response to the query. This process measures the end-to-end utility of our retrieval quality in generating accurate answers. 
\end{itemize}  

\noindent  
While retrieval effectiveness remains our primary focus, we also evaluate QA accuracy to understand its downstream impact. This dual evaluation serves two main purposes. First, comprehensively retrieving all relevant evidence provides the necessary context for downstream tasks; conversely, overlooking critical documents can significantly degrade QA performance. If retrieval performance and QA accuracy exhibit similar trends in experiments, it would indicate that the retrieval stage is a decisive factor in both constraining and improving final outcomes. Since tasks requiring multi-step reasoning inherently involve synthesizing information from multiple sources, exhaustive retrieval is indispensable. Consequently, we track both retrieval effectiveness and QA accuracy to provide a comprehensive assessment of the system’s overall utility.

\paragraph{Multi-hop QA Datasets.}
To evaluate the proposed system’s capabilities for complex multi-step reasoning,  we utilize the following multi-hop QA benchmarks that necessitate multi-step reasoning across multiple documents:
\begin{itemize}
\item \textbf{HotpotQA} \cite{yang2018hotpotqadatasetdiverseexplainable}: Requiring reasoning over entity-linked passages with distractors. (Used \textit{dev\_fullwiki\_v1} split)

\item \textbf{2WikiMultiHopQA} \cite{ho2020constructingmultihopqadataset}:  A multi-hop QA dataset explicitly designed to eliminate shortcut reasoning and require cross-document inference. (Used \textit{dev} split)

\item \textbf{MuSiQue} \cite{trivedi2022musiquemultihopquestionssinglehop}: Curated chains of up to 4 hops with no redundant context, demanding implicit bridging across passages. (Used \textit{ans\_v1.0\_dev} split)
\end{itemize}
\textbf{MuSiQue} presents the highest intrinsic complexity due to its extended reasoning depth. For example, a question like "What is the nationality of the author who influenced the philosopher cited in [Event X]?" requires aggregating information across four distinct documents. In contrast,  \textbf{2WikiMultiHopQA} emphasizes error sensitivity: Its strict dependency structure causes errors from early retrieval stages to propagate through reasoning, leading to reduced accuracy compared to \textbf{HotpotQA}. This structural rigidity makes it particularly challenging for cascade-based systems.

\paragraph{Single-hop QA Datasets:}
\begin{itemize}
\item \textbf{Natural Questions} \cite{kwiatkowski-etal-2019-natural}: Evaluates the ability to directly retrieve and extract information from a single document based on real user queries. (Used \textit{text-v1.0-simplified-nq-dev-all} split.)

\item \textbf{SQuAD} \cite{rajpurkar2016squad100000questionsmachine}: Assesses fundamental contextual understanding and retrieval skills through reading comprehension tasks that require the extraction of answer spans from documents. (Used \textit{val} split.)
\item \textbf{TriviaQA} \cite{joshi2017triviaqalargescaledistantly}: Measures straightforward information retrieval and extraction capabilities using a large-scale question-answer dataset. (Used \textit{unfiltered-web-dev} split.)
\end{itemize}

\noindent
Single-hop QA datasets serve to verify that the proposed system performs reliably on straightforward retrieval. By evaluating tasks across varying difficulty levels, we aim to assess whether the system can adapt to both simple and complex scenarios without overfitting to more challenging tasks.

For all datasets, we randomly sampled 200 questions as the
evaluation set.

\subsubsection{Retrieval Configuration}
We use BM-25 as the retrieval model for all experiments. All agents utilize the same model with identical configurations. The retrieval depth is fixed at the top 10 results, and documents are processed at the paragraph level. For \textbf{Natural Questions}, paragraphs are split into chunks of 256 tokens to accommodate its longer context requirements.

\subsubsection{Agent Configurations and Baselines}

\paragraph{System Configurations.}
We evaluate the following configurations, including a baseline and proposed systems:
\begin{itemize}
    \item \textbf{One-step Retrieval (Naive-RAG, Baseline)}: A single-step retrieval model without iterative refinement, representing a standard Retrieval-Augmented Generation approach.
    \item \textbf{Proposed Single-Agent System}: Executes a knowledge-aware  retrieval algorithm through a single agent, as detailed in Algorithm~\ref{alg:agent_inference}.
    \item \textbf{Proposed Multi-Agent System}: A competition-based extension, where up to three agents iteratively refine queries and share context through dynamic knowledge updates.
\end{itemize}

\paragraph{Language Models for Agents.}
Each agent uses \texttt{GPT-4o-mini} \cite{openai2023gpt4o-mini} by default. For \textbf{2WikiMultiHopQA} and \textbf{MuSiQue} that present higher complexity, we additionally test \texttt{GPT-4o} \cite{openai2024gpt4o} to assess performance scalability.

\paragraph{QA Module.}
A separate QA module generates answers from retrieved contexts in a standard RAG setting. By default, we use \texttt{GPT-3.5-turbo-0125} \cite{openai2023gpt35turbo}. For \textbf{2WikiMultiHopQA} and \textbf{MuSiQue}, we additionally test \texttt{GPT-4o-mini} to isolate the impact of the quality of the generation.

\subsubsection{Evaluation Metrics}
We measure two complementary aspects:
\begin{itemize}
    \item \textbf{Retrieval F1}: Token-level overlap (F1 score) between retrieved and ground-truth supporting documents, reflecting evidence completeness.
    \item \textbf{Answer F1}: Token-level overlap (F1 score) between predicted and ground-truth answers, indicating downstream task utility.
\end{itemize}

\subsubsection{Implementation.}
The agent systems are implemented in Python using \texttt{LangGraph} \cite{langgraph2023}, a library for building stateful, multi-agent workflows, where each agent can perform retrieval or reasoning with a cache such as knowledge, context, or query history.
\subsection{Results}
\label{subsec:main_results}

\begin{figure*}[!t]
    \centering
    \begin{subfigure}{\textwidth}
        \centering
        \includegraphics[width=0.8\textwidth]{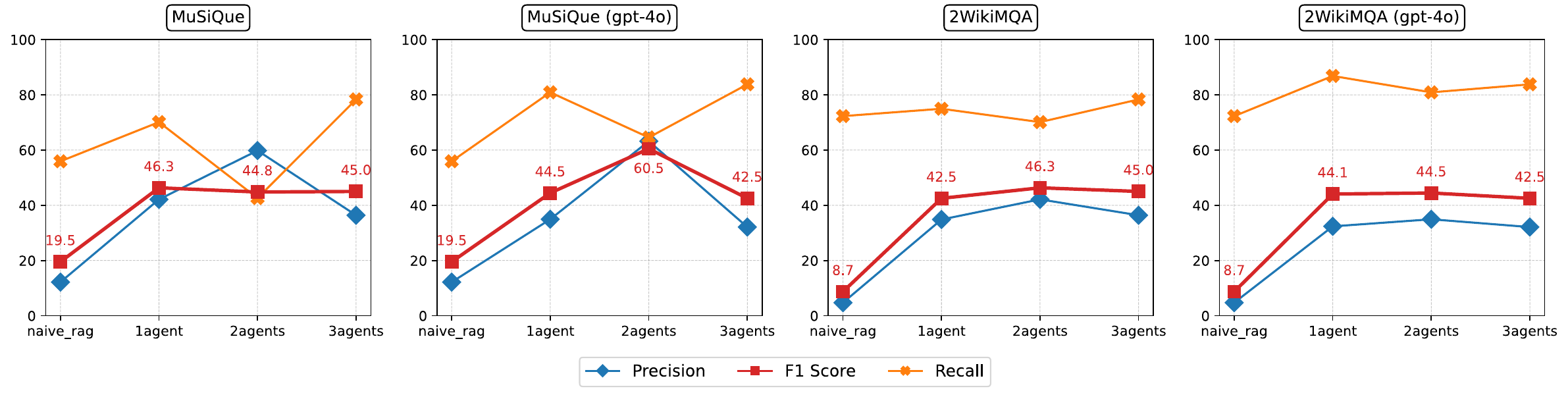}
    \end{subfigure}
    \vspace{1em}
    \begin{subfigure}{\textwidth}
        \centering
        \includegraphics[width=0.8\textwidth]{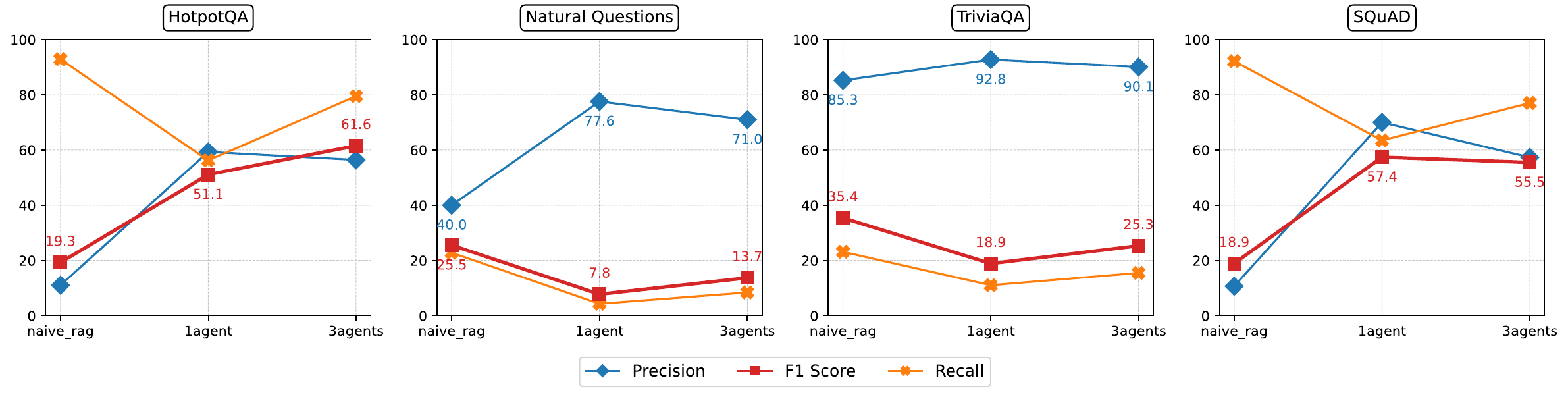}
    \end{subfigure}
    \caption{Comparison of retrieval F1 across different datasets and agent configurations. The top sub-figure focuses on MuSiQue and 2WikiMultiHopQA, while the bottom sub-figure illustrates HotpotQA, TriviaQA, Natural Questions, and SQuAD.}
    \label{figure:comparison}
\end{figure*}

\begin{figure*}[!t]
  \centering
  \includegraphics[width=\linewidth]{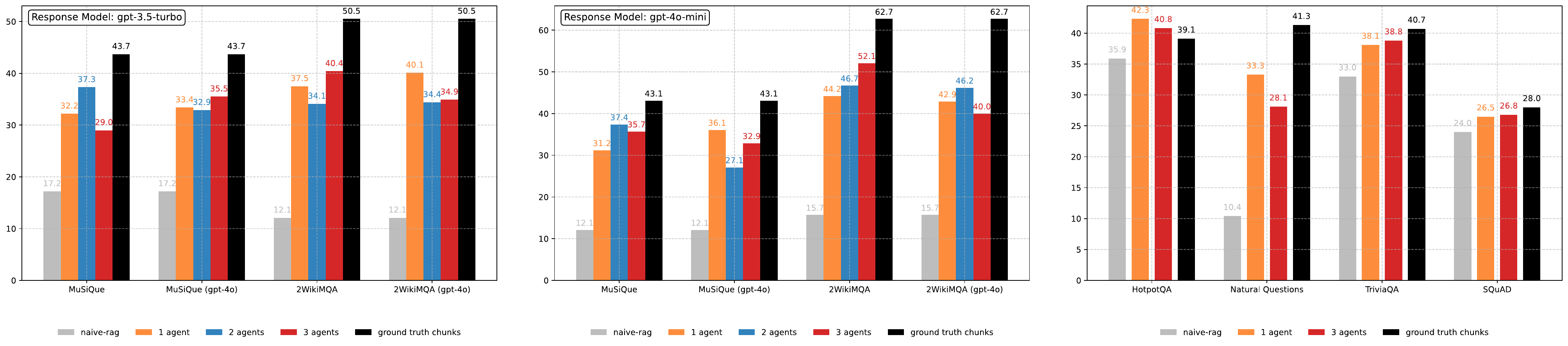}
  \caption{Answer F1 scores across retrieval systems and agent configurations. Multi-agent systems often yield higher F1 on complex datasets due to more precise context retrieval.}
  \label{figure:comparison_general}
\end{figure*}

Figures~\ref{figure:comparison} and \ref{figure:comparison_general} illustrate \textbf{retrieval F1} and \textbf{answer F1} scores, respectively, across all experimental datasets. The following subsections analyze these results in depth.

\subsubsection{Effectiveness of the Agent System}
\begin{itemize}
\item \textbf{Both single- and multi-agent systems outperform one-step retrieval.} 
We observe consistent gains in retrieval effectiveness and QA accuracy across a variety of datasets, compared to one-step retrieval.

\item \textbf{The agent systems excel particularly in challenging tasks.}
In \textbf{MuSiQue} and \textbf{2WikiMultiHopQA}, performance improvements are especially pronounced,
with multi-agent systems exhibiting a significant boost over single-agent systems.
This appears to stem from agents revising queries from different perspectives, thereby incorporating the necessary intermediate information.

\item \textbf{Agent systems can surpass ground-truth evidence in certain scenarios.} For instance, in \textbf{HotpotQA}, both single- and multi-agent systems outperform the ground-truth evidence.
We attribute this result to the agent's structured problem-solving approach and its feedback loop for self-correction.
Even a single-agent system continuously refines its knowledge structure by identifying and filling information gaps,
thereby augmenting or correcting what human-annotated evidence might overlook.
\end{itemize} 
\noindent
We also note that the structural characteristics of a dataset can affect the performance of agent systems.
For instance, in a dataset like \textbf{Natural Questions}, where most answers are concentrated within a single document, uemploying an excessive number of agents can actually increase noise and lower precision.
Conversely, even with a dataset like \textbf{2WikiMultiHopQA}, which requires referencing multiple documents, if there is train–test overlap or certain shortcut elements in the design, an agent can solve problems without genuine multi-step reasoning, limiting potential performance gains.

\begin{figure*}[!t]
  \centering
  \includegraphics[width=\linewidth]{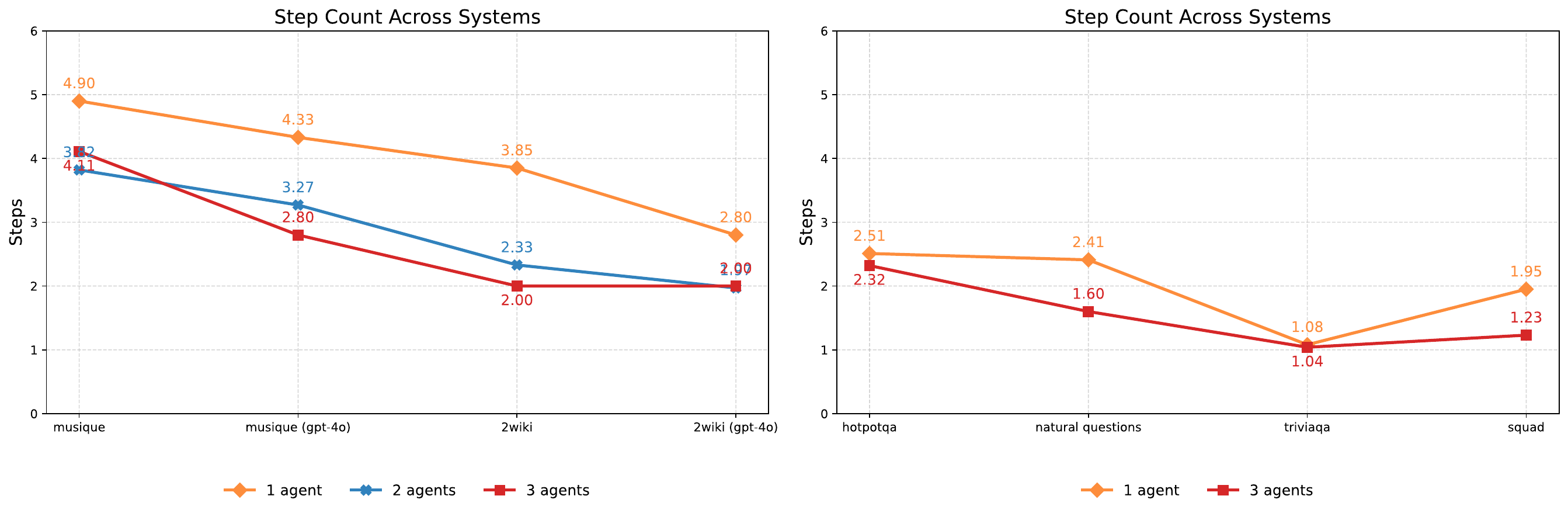}
  \caption{Convergence behavior of the Knowledge-Aware Agent Retrieval Algorithm. The figure presents the number of retrieval iterations required across different agent configurations, evaluating the impact of iterative refinement on retrieval efficiency. The results highlight how agent collaboration influences convergence speed, demonstrating the effectiveness of knowledge-aware retrieval.}
  \label{fig:convergence_behavior}
\end{figure*}

\subsubsection{Multi-Agent Extension}
\label{sec:multi_agent_breakdown}

In this section, we aim to answer the following questions:
\begin{itemize}
    \item What is the optimal number of agents?
    \item Does the benefit of a multi-agent system over a single-agent system vary depending on task characteristics?
    \item Are multiple agents consistently advantageous, or do external factors  moderate the benefits of multi-agent systems?
\end{itemize}

We focus on two multi-hop QA datasets, \textbf{MuSiQue} and \textbf{2WikiMultiHopQA}, which serve as the most challenging tasks and highlight the performance gaps between single-agent and multi-agent systems. To this end, we vary: 
(i) the number of agents, 
(ii) the LLM employed by the agents, and 
(iii) the generator (reader LLM),
to investigate their effects.

\paragraph{Agent Scaling.} 
We scale the number of agents up to three and observe the following:

\textbf{Retrieval Performance.}
Increasing the agent count does not yield strictly linear gains, as results depend on the dataset and the specific LLM configurations.
For instance, regardless of configuration, two agents generally achieve higher precision than a single agent, yet three agents do not necessarily surpass two agents in most configurations.

\textbf{Downstream QA Performance.}  
A similar pattern is observed in downstream QA tasks, where performance improvements do not always scale linearly with the number of agents. For instance, on \textbf{MuSiQue}, a multi-agent system configured with \texttt{GPT-4o-mini} surpasses the single-agent baseline; however, when configured with \texttt{GPT-4o}, the single-agent system can sometimes yield superior results, depending on the generator LLM. These discrepancies underscore that while multi-agent cooperation  appears to be beneficial in challenging tasks, simply increasing the agent count does not guarantee consistent performance gains.

\paragraph{Effect of Agent LLM Configuration.}
We next examine how changes in the LLM used to configure the agents impact both retrieval and downstream QA.

\textbf{Retrieval Performance.}  
Overall, the two-agent system outperforms the single-agent system across datasets, regardless of the agent LLM configuration. In contrast, the three-agent system tends to primarily boost recall, and this trend appears to be independent of which LLM is used to configure them.

\textbf{Downstream QA Performance.}  
When considering multi-agent systems as a whole, including both two-agent and three-agent setups, they generally achieve superior performance compared to the single-agent system, winning in six out of eight configuration combinations. However, the relative performance between the two-agent and three-agent systems depends on the specific LLM configuration. On the \textbf{MuSiQue} dataset, for instance, the highest efficiency (37.3/43.7) is obtained when the agents are configured with GPT-4o-mini, surpassing other configurations while approaching the performance achieved with ground truth chunks.

\paragraph{Effect of the Generator LLM.}
The generator produces final answers from the last context returned by the agent systems, using a distinct LLM. We examine the impact of this choice on downstream QA performance, organized by dataset.

\textbf{MuSiQue.}
In most settings, multi-agent systems outperform the single-agent baseline. An exception arises when the single-agent system is configured with \texttt{GPT-4o} and uses \texttt{gpt-3.5-turbo} as the generator, briefly surpassing certain multi-agent setups. Meanwhile, using ground truth chunks for retrieval yields similar accuracy (43.7/43.1), regardless of which generator LLM is employed. When agent systems handle retrieval, a two-agent system achieves 37.3/37.4 with different generators, indicating the highest performance among agent-based configurations.

\textbf{2WikiMultiHopQA.}  
Similarly, multi-agent systems generally surpass the single-agent baseline. One exception arises when a single-agent system configured with \texttt{GPT-4o} and using \texttt{gpt-3.5-turbo} as the generator outperforms certain multi-agent configurations. In this dataset, the accuracy of ground truth retrieval differs markedly depending on the generator LLM (50.5/62.7), and the three-agent system achieves the highest efficiency (40.4/52.1) among agent-based methods.

\paragraph{Overall Findings and Takeaways.}
\begin{itemize}
    \item \textbf{Two-agent systems optimize cost-performance tradeoffs}: Compared to single-agent setups, two-agent systems achieve the highest retrieval effectiveness and downstream QA accuracy, whereas adding a third agent often offers only marginal gains and may even degrade performance in certain scenarios.
    
    \item \textbf{Saturation in retrieval coverage}: Retrieval F1 levels off with two agents, indicating that our method does not indiscriminately add context but rather precisely detects information gaps. This effect appears to be most clearly manifested in the two-agent configuration.
    
    \item \textbf{Smaller LLMs may suffice for coordination}: Agents configured with \texttt{GPT-4o-mini} often exhibit performance comparable to their heavier counterpart, \texttt{GPT-4o}. In two-agent systems, the efficiency of \texttt{GPT-4o-mini}-configured agents is particularly notable across both retrieval and downstream QA tasks. Effective collaboration relies more on structured interaction protocols than on the raw capabilities of the LLM powering the agents.
    
    \item \textbf{Retrieval gains often translate into downstream QA improvements}:  
    Multi-agent systems generally outperform single-agent setups in retrieval, and this advantage tends to carry over to QA tasks.  
    However, the optimal configuration, whether two-agent or three-agent, can vary depending on the specific scenario and generator employed.
    
    \item \textbf{Architectural advantages over sequential methods}:
    Although \texttt{GPT-4o} is generally recognized for its superior reasoning capabilities compared to \texttt{GPT-4o-mini},
    the strong performance that \texttt{GPT-4o-mini} exhibits in multi-agent systems underscores the intrinsic value of our architecture.
    Substantial improvements in retrieval and downstream QA highlight its effectiveness,
    while the diminishing returns observed in three-agent systems suggest that scaling beyond a certain point may hit an architectural limit to further optimization.

\end{itemize}

\subsubsection{Comparison with Single-Step Baseline}  
\label{sec:comparison_rag_groundtruth} 
In Section~\ref{sec:multi_agent_breakdown}, we demonstrated that multi-step retrieval improves precision on complex tasks. Here, we analyze the limitations of a single-agent retrieval system and its impact on recall.

Compared to a single-step baseline, our agent-based retrieval system shows a recall reduction in datasets like \textbf{Natural Questions} and \textbf{TriviaQA}. This is because, at the search depth used in our retrieval, one-step retrieval often provides sufficiently high recall. For instance, except for chunked Natural Questions, \textbf{MuSiQue} also achieves a recall above .6 with a single retrieval step.

This recall advantage in one-step retrieval stems from the evaluation metric, which considers all ground-truth documents as correct. However, despite its high recall, naive retrieval methods lack filtering mechanisms, leading to unrefined document selection and reduced accuracy in downstream QA, particularly in tasks requiring precise cross-evidence synthesis.
 
\subsubsection{Comparison with Iterative Retrieval Methods}  
Recent studies on multi-step retrieval \cite{trivedi2023interleavingretrievalchainofthoughtreasoning} proposed a chain-of-thought (CoT) retrieval method, in which the LLM generates a CoT from the retrieval results and appends it to the query in subsequent iterations. The method reports a recall of 57.1 and an answer F1 of 36.3 on the Musique dataset using a state-of-the-art model, and a recall of 90.7 and an answer F1 of 68.0 on the 2WikiMultiHopQA dataset. However, precision metrics were not reported, limiting direct comparison.

In contrast, the proposed method achieves a recall of 42.61, an F1 of 44.81, and a precision of 59.81 on the Musique dataset. The answer F1 score is 37.3. On the 2Wiki dataset, the method achieves a recall of 78.3, an F1 of 45.01, and a precision of 36.39, with an answer F1 score of 40.4.

The proposed system employs a targeted query refinement strategy that explicitly identifies missing information at each reasoning step. By pinpointing critical unresolved evidence and formulating concise follow-up queries targeting these gaps, the process ensures each retrieval iteration addresses a well-defined sub-question. This reduces unnecessary noise while enhancing interpretability. On the Musique dataset (up to 4 hops), the gap identification strategy effectively eliminates noise and focuses on essential evidence, achieving balanced recall (42.61) and precision (59.81).  

The high recall of interleaving CoT on 2WikiMultiHopQA is attributed to the dataset’s structure: questions are based on Wikidata triples, where both questions and evidence chains follow explicit relational paths. Consequently, CoT-augmented queries align precisely with the intended evidence documents, driving high recall. 

In conclusion, the systematic gap identification strategy, enabled by the agent system’s capacity to track and prioritize unresolved evidence across reasoning steps, indicates increasing effectiveness as task complexity escalates compared to methods reliant on unstructured query expansion.

\subsubsection{Efficiency Analysis}
\label{subsubsec:efficiency_analysis}
\paragraph{Observations on Efficiency:}
Figure~\ref{fig:convergence_behavior} presents the convergence behavior across agent configurations. The efficiency trends across agent configurations provide insights into their relative effectiveness in different tasks.

\begin{itemize}
\item \textbf{Three-Agent Setups Improve Convergence but Level Off on Simple Tasks}: Systems with 3 agents require fewer iterations to converge compared to single-agent setups, e.g., 2.8 steps vs. 4.9 steps on Musique (LLM: \texttt{GPT-4o}). However, this reduction levels off for simpler tasks like TriviaQA (1.04 steps for 3 agents vs. 1.08 for 1 agent).
\item \textbf{Multi-Agent Systems Exhibit Consistent Efficiency on Complex Tasks}: Multi-agent systems show consistent step counts on complex tasks (e.g., 2.00 steps for 3 agents on 2WikiMultiHopQA) but minimal gains on single-hop tasks like SQuAD (1.23 steps vs. 1.95 for 1 agent).
\item \textbf{Two-Agent Systems Offer the Best Accuracy-Efficiency Trade-Off}: The 2-agent configuration achieves the strongest task-agnostic performance, with retrieval F1 of 46.3 on Musique (LLM: \texttt{GPT-4o-mini}) at 3.82 steps, demonstrating robust adaptation to both multi-hop and simpler QA formats.
\item \textbf{Task Complexity Determines the Optimal Agent Count}: These findings suggest that multi-agent configurations are particularly beneficial for complex, multi-hop tasks, where reducing the number of iterations can lead to efficiency gains. For simpler, single-hop tasks, the marginal benefits of additional agents diminish, indicating that lower agent counts may be more optimal to avoid unnecessary overhead.
\end{itemize}


\begin{table}[t]
\centering
\caption{Representative Efficiency Analysis on MuSiQue.}
\begin{tabular}{lcccc}
\toprule
\textbf{Model} & \textbf{Agents} & \textbf{Steps} & \textbf{Cost Units} & \textbf{Retrieval F1}\\
\midrule
GPT-4o-mini & 1-agent\phantom{s} & 4.90 & 4.90 & 46.3 \\
GPT-4o-mini & 2-agents & 3.82 & 7.64 & 44.8 \\
GPT-4o-mini & 3-agents & 4.11 & 12.33 & 45.0 \\ 
GPT-4o & 1-agent\phantom{s} & 4.33 & 43.3 & 44.5 \\
GPT-4o & 2-agents & 3.27 & 65.4 & 60.5 \\
GPT-4o & 3-agents & 2.80 & 84.0 & 42.5 \\
\bottomrule
\end{tabular}
\label{tab:efficiency_musique}
\end{table}

\begin{table}[t]
\centering
\caption{Representative Efficiency Analysis on 2WikiMultiHopQA.}
\begin{tabular}{lcccc}
\toprule
\textbf{Model} & \textbf{Agents} & \textbf{Steps} & \textbf{Cost Units} & \textbf{Retrieval F1}\\
\midrule
GPT-4o-mini & 1-agent\phantom{s} & 3.85 & 3.85 & 42.5 \\
GPT-4o-mini & 2-agents & 2.33 & 4.66 & 46.3 \\
GPT-4o-mini & 3-agents & 2.99 & 8.97 & 45.0 \\
GPT-4o & 1-agent\phantom{s} & 2.80 & 28.0 & 44.1 \\
GPT-4o & 2-agents & 2.00 & 40.0 & 44.5 \\
GPT-4o & 3-agents & 2.00 & 60.0 & 42.5 \\
\bottomrule
\end{tabular}
\label{tab:efficiency_2wiki}
\end{table}

\paragraph{Cost-Benefit Analysis}

Tables~\ref{tab:efficiency_musique} and \ref{tab:efficiency_2wiki} detail computational costs, 
where cost units are calculated as
\[
\text{Cost Units} = (\text{Steps}) \times (\text{Agents}) \times (\text{Model Cost Factor}),
\]
based on an approximate 10:1 cost ratio between \texttt{GPT-4o} and \texttt{GPT-4o-mini} API calls (estimated from API pricing benchmarks).

While configurations employing \texttt{GPT-4o} achieve marginally higher retrieval F1 scores in specific setups, their order-of-magnitude cost increase often renders these gains impractical. For instance, the 3-agent system configured with \texttt{GPT-4o} incurs 84.0 cost units for 42.5 retrieval F1 on MuSiQue, whereas the 2-agent system configured with \texttt{GPT-4o-mini} attains comparable performance (45.0 retrieval F1) at 12.33 cost units. Given the marginal performance differences, systems configured with \texttt{GPT-4o-mini} offer a cost-efficient balance for scenarios that demand high computational efficiency.


\subsubsection{Qualitative Analysis: Query Planning Examples}  
The following case study illustrates how a three-agent system, operating under a competition-based scheme, processes a 4-hop question from the MuSiQue dataset.

Each agent independently manages two dynamic knowledge states: \emph{What is Known}, which consists of facts verified through the agent’s independent retrieval process and \emph{What is Required}, which includes unresolved sub-queries specific to the agent’s reasoning path. These states evolve autonomously until at least one agent resolves all requirements (\texttt{None} status), triggering termination.

\begin{figure}[t]
\centering
\scriptsize
\begin{verbatim}
+------------------------------------------------+
| Process Flow Diagram                           |
+------------------------------------------------+
User Query: "Who brought Hokkien to the country on 
the natural boundary between the tournament host 
and the country where A Don is from?"

Iteration 1:
  Suggester1: 
    - Known: Hokkien spread by Han immigrants from 
      southern Fujian.
    - Required: Identify the natural boundary country.
  Suggester2:
    - Known: Major spread regions include Myanmar, 
      Indonesia, Malaysia, etc.
    - Required: Identify host and A Don's countries.
  Suggester3:
    - Known: Primary spread group: Hoklo immigrants.
    - Required: Specify the target country.

Iteration 2:
  Suggester1:
    - Suggested Myanmar.
    - Required: Identify the specific local group.
  Suggester2:
    - Considered Spain as host; A Don from Laos.
    - Required: Confirm Myanmar's role.
  Suggester3:
    - Confirmed Myanmar borders Laos.
    - Required: Identify a propagating group in Myanmar.

Iteration 3:
  Suggester1:
    - Concluded spread by Hoklo immigrants in Myanmar.
    - Required: Detail the demographic.
  Suggester2:
    - Confirmed the Myanmar–Laos border.
    - Required: Specify the transmission pathway.
  Suggester3:
    - Identified a major Han community.
    - Required: Verify specific group identity.

Iteration 4:
  Suggester1:
    - Verified historical use in Myanmar.
    - Required: Specify a time period or subgroup.
  Suggester2:
    - Confirmed 19th-century immigration.
    - Required: Obtain documentary evidence.
  Suggester3:
    - Found "Han Chinese emigrants" in a document.
    - Required: Acquire additional details.

Iteration 5:
  Suggester1:
    - Sought further literature.
    - Required: Identify a specific subgroup.
  Suggester2:
    - Reviewed historical records.
    - Required: Specify a particular family or group.
  Suggester3:
    - Directly confirmed "Han Chinese emigrants".
    - Required: None

Iteration 6 (Termination):
  Process terminates as Suggester3's Required state is None.
\end{verbatim}
\caption{Condensed process flow diagram showing the evolution of knowledge across iterations in the multi-agent dialogue.}
\label{fig:multi-agent-dialogue-excerpt}
\end{figure}

Agents initially possessed only basic facts (e.g., that Hokkien was spread by Han immigrants from southern Fujian) without the crucial details required by the query, such as the natural boundary country or the specific immigrant group. Through iterative dialogue, the system progressively refined this information. This case study exemplifies the systematic approach taken by agents to disambiguate and refine knowledge through controlled iterations. The following key aspects highlight the structured nature of the multi-agent reasoning process.

\begin{itemize}
    \item \textbf{Incremental Knowledge Expansion:}  
    Early iterations provided only rudimentary facts. In Iterations 2–3, the natural boundary country (Myanmar) was identified and it was confirmed that Hoklo immigrants spread Hokkien. Iterations 4–5 further clarified the presence of a major Han community in Myanmar and introduced historical context (including 19th-century Chinese migration), culminating in Iteration 6 with the direct confirmation of “Han Chinese emigrants.”
    
    \item \textbf{Role Differentiation Among Agents:}  
    Distinct roles emerged among the agents: Suggester1 focused on geographic relationships (e.g., clarifying Myanmar–Laos connections), Suggester2 concentrated on identifying contextual details (e.g., determining the host and A Don's countries, and confirming the boundary), and Suggester3 specialized in consolidating historical evidence to pinpoint the propagating group.
    
    \item \textbf{Adaptive Query Planning:}  
    Agents evolved their queries from broad prompts (e.g., "Which country forms the natural boundary?") to more specific inquiries (e.g., detailing the transmission pathway and verifying group identity), systematically reducing ambiguity.
    
    \item \textbf{Convergence on a Final Answer:}  
    The iterative refinement led to the resolution of outstanding gaps; by Iteration 6, when Suggester3’s \emph{Required} state reached \texttt{None}, the process terminated with a precise, synthesized answer.
\end{itemize}

Overall, the multi-agent dialogue demonstrates how iterative knowledge updates, adaptive query planning, and complementary agent roles enable the system to progressively refine incomplete information into a coherent answer. This process effectively illustrates the system's capability to resolve complex 4-hop questions through structured collaboration.

\section{Conclusions}
We propose \textbf{Knowledge-Aware Iterative Retrieval for Agent Systems}, a novel agent framework designed to optimize iterative information retrieval. Multi-step reasoning problems pose a critical challenge by requiring the efficient connection of fragmented information across diverse knowledge sources, a challenge that it addresses by dynamically refining search queries through a human-like cognitive cycle. The proposed approach, driven by agent-based autonomous decision-making, systematically filters and updates retrieved contexts using knowledge-based relevance criteria, ensuring both informational diversity (through exploration of multiple reasoning paths) and accuracy (by pruning noise through shared or agent-specific caches).

A key novelty of the proposed system is its systematic gap detection and targeted re-querying mechanisms. Compared to iterative retrieval approaches that rely on static search strategies, it systematically detects unresolved gaps and directs subsequent searches to target these specific deficits. The system's modular design supports flexible scaling from single-agent to multi-agent configurations; in multi-agent collaboration, knowledge can be shared or refined contexts can be flexibly integrated based on accumulated information, thereby enhancing overall retrieval performance while maintaining computational efficiency through lightweight models.

Comprehensive experiments on six open-domain QA datasets, including multi-step reasoning tasks, demonstrate that the proposed system outperforms single-step retrieval baselines. Both retrieval performance and downstream QA accuracy were taken into account. As task complexity increases, compared to iterative retrieval approaches that rely on static search strategies, its systematic gap detection and targeted re-querying mechanisms achieve a superior balance between precision and recall in the search context. Similarly, the multi-agent configuration exhibits a marked advantage over single-agent setups under more complex tasks. An in-depth analysis of agent scaling reveals that a cost-efficient configuration based on agents with a \texttt{GPT-4o-mini} setup attains performance comparable to or even exceeding that of heavier counterparts (e.g., \texttt{GPT-4o} in our experiments). These findings indicate that heavier counterparts are not necessarily optimal for effective agent collaboration. Instead, improvements in reasoning capabilities for tasks that require the integration of diverse information are best achieved through optimal multi-agent collaboration, suggesting significant potential for cost-efficient and scalable solutions.

\section{Limitations}
We explored only a competition-based model in our multi-agent extension. Although we also proposed a knowledge-sharing collaboration structure, we limited our experiments to the competition-based communication model due to its implementation simplicity. Our experimental results indicate that improvements in reasoning capabilities are achieved through the independent agent collaboration structure, and they suggest the existence of an optimal collaboration framework. Future work could investigate alternative models, such as non-competitive or hybrid collaboration structures, to further optimize multi-agent synergy.

In our experiments, simple prompts were employed to mitigate prompt sensitivity, and the system maintained a negligible failure rate. However, as collaborative elements in agent communication strengthen and role differentiation emerges, structured messaging that ensures consistent adherence to protocols across agents will become increasingly important to reduce misinterpretation risks. In such cases, automating prompt generation during planning may be considered to further streamline message exchange and foster resilient collaboration in multi-agent systems. 

Since the study primarily focuses on open-domain tasks, further validation on domain-specific tasks would be beneficial in fully assessing the generalizability of the findings.

Open-source LLMs were not evaluated in this study because many such models struggle with structured collaboration (e.g., \cite{liu2023agentbenchevaluatingllmsagents}), despite their potential benefits in terms of accessibility and scalability.


\end{document}